%% file: 00_irmv_ant.tex
\documentclass[lettersize,journal]{IEEEtran}
\usepackage{amsmath,amsfonts}
\usepackage{algorithmic}
\usepackage{algorithm}
\usepackage{array}
\usepackage[caption=false,font=normalsize,labelfont=sf,textfont=sf]{subfig}
\usepackage{textcomp}
\usepackage{stfloats}
\usepackage{url}
\usepackage{verbatim}
\usepackage{graphicx}
\usepackage{cite}
\usepackage{booktabs}
\usepackage{float}
\usepackage{censor} 

\renewcommand{\censor}[1]{#1}

\begin{document}

\title{High-Degree-of-Freedom Lightweight Bioinspired Leg for Enhanced Mobility in Small Robots}

\author{\censor{Haoqi Han}, \censor{Yifei Yu}, \censor{Jiaming Zhang}, \censor{Xinru Cui}, \censor{Linxi Feng}, \censor{Hesheng Wang,~\IEEEmembership{Senior Member,~IEEE}}
\thanks{
        \blackout{This work was supported in part by the Natural Science Foundation of China under Grant 62225309, U24A20278, 62361166632, U21A20480 and 62403311. (Corresponding Author: Hesheng Wang).}

        \blackout{Haoqi Han is with the Department of Computer Science and Engineering, Shanghai Jiao Tong University, Shanghai, 200240, China (e-mail: hhq123@sjtu.edu.cn).}

        \blackout{Yifei Yu is with the Department of Automation Engineering,Shanghai University of Electric Power,Shanghai,China(e-mail:yuyifei@mail.shiep.edu.cn).}
        
        \blackout{Jiaming Zhang is with the Department of Naval Architecture and Marine Engineering, Shanghai Jiao Tong University, Shanghai, 200240, China (e-mail: zjm4687@sjtu.edu.cn).}
        
        \blackout{Xinru Cui, Linxi Feng are with the Department of Automation, Shanghai Jiao Tong University, Shanghai, 200240, China (e-mail: ).}

        \blackout{Hesheng Wang is with the Department of Automation, Key Laboratory of System Control and Information Processing of Ministry of Education, Key Laboratory of Marine Intelligent Equipment and System of Ministry of Education, Shanghai Engineering Research Center of Intelligent Control and Management, Shanghai Jiao Tong University, Shanghai, 200240, China (e-mail: wanghesheng@sjtu.edu.cn).}
}
\thanks{Manuscript received April 19, 2021; revised August 16, 2021.}}

\markboth{Journal of \LaTeX\ Class Files,~Vol.~14, No.~8, August~2021}%
{Shell \MakeLowercase{\textit{et al.}}: A Sample Article Using IEEEtran.cls for IEEE Journals}

\IEEEpubid{0000--0000/00\$00.00~\copyright~2021 IEEE}

\maketitle

\begin{abstract}
In microrobotics, enhancing locomotion capabilities by increasing the degrees of freedom (DoF) of leg mechanisms under severe spatial constraints remains a significant challenge.
Inspired by insect locomotion, this paper presents a novel micro-scale parallel leg mechanism with four degrees of freedom, and systematically analyzes its mechanical design, electrical system, and kinematics.
The design incorporates two spherical five-bar linkages to achieve spatial motion within a parallel four-bar configuration.
Furthermore, a concentric design strategy is employed to simplify the analytical solution of the leg kinematics.
Due to the parallel system architecture, all actuators are located on the main body, substantially reducing the equivalent inertia of moving parts compared to traditional high-DOF leg structures.
The total mass of the system is only 18.9 g, with an end-effector output force of approximately 0.5 N and a workspace exceeding 22255 mm³.
Experimental results demonstrate that the proposed single-leg mechanism achieves excellent motion flexibility, highlighting its potential for micro bio-inspired robotics.

\end{abstract}


\input{01_introduction}
\input{02_design}

\input{03_analysis}

\input{04_experiment}
\input{05_conclusion}
\input{06_reference}

\end{document}

%% file: 01_introduction.tex
\section{Introduction}

\begin{figure}
    \centering
    \includegraphics[width=1.0\linewidth]{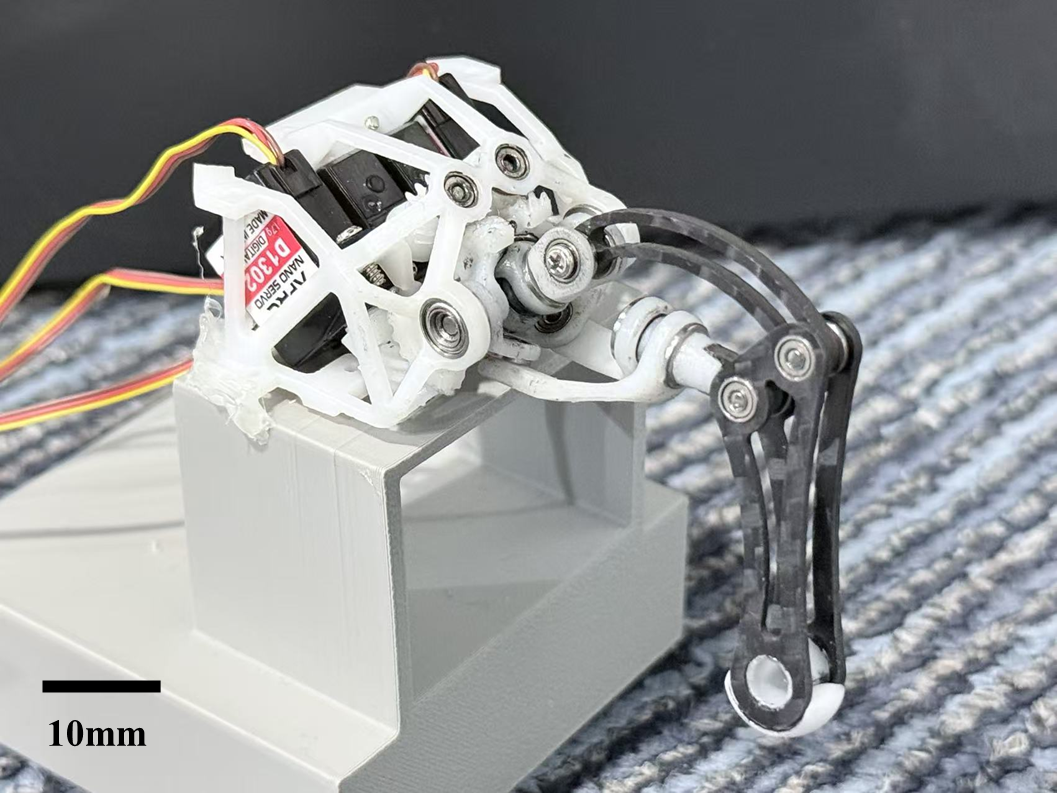}
    \caption{A novel micro-scale parallel leg mechanism}
    \label{fig:ant_real}
\end{figure}

In recent years, small-scale legged robots often referred to as rat-sized or insect-scale platforms have garnered significant attention \cite{yang_grand_2018}.
Due to their compact physical dimensions, these robots possess a unique advantage in navigating unstructured and confined environments that are inaccessible to larger machines\cite{benzi2022whole,8391705}.
Consequently, they are envisioned as ideal candidates for critical tasks such as post-disaster search and rescue \cite{kohut_medic_2011}, environmental reconnaissance, and material transport in narrow, hazardous zones.
However, endowing micro-robots with biological-level agility and payload capacity under extreme size constraints remains a formidable challenge.

The core challenge in developing high-performance micro-robots lies in the inherent trade-off between actuator limitations and mechanical structure.
Due to constraints in micro-electromechanical systems (MEMS), miniature actuators typically exhibit low power density and limited output torque \cite{9274422,lei2025design,fujita2002microactuators}.
To adapt to these constraints and reduce system weight, existing studies often compromise on kinematic design by reducing degrees of freedom (DoF).
Currently, mainstream small multi-legged robots mostly employ single-leg designs with fewer than three DoFs \cite{hollar2003solar,zhang2025bridging,10122688}, which, while simplifying control, severely limits omnidirectional mobility and posture adjustment in complex terrains.

Conversely, designs aiming for higher DoF often rely on conventional serial mechanisms, with actuators mounted directly at the joints.
This serial configuration significantly increases the oscillating mass of the leg, leading to high rotational inertia that severely limits dynamic response and agility.
Although cutting-edge research—such as the Harvard RoboBee and HAMR series—has demonstrated high-frequency motion at the insect scale using piezoelectric actuators \cite{ma_controlled_2013, baisch_high_2014, jayaram_scaling_2020}, as well as soft artificial muscles \cite{chen_controlled_2019}, electromagnetic servo motors remain a more robust option for applications that demand high payload capacity and stable interaction.
Therefore, a new structural paradigm is needed—one that resolves the trade-off between high DoF and low leg inertia while maintaining integrated servo actuation.

Nature offers an elegant solution to this trade-off.
These DoF enable organisms to control their limbs flexibly in complex environments.
The limbs of many organisms, though seemingly simple linkage systems, exhibit high kinematic redundancy—often possessing up to seven degrees of freedom (DoF) \cite{dickinson_wing_1999,6974009,7339708}.
Crucially, unlike serial robots, arthropods employ a proximal actuation strategy: although their legs consist of serially connected rigid exoskeletons, precise motion is achieved through a parallel muscle system located within the thorax \cite{wood_microrobot_2008}.
This configuration minimizes the distal inertia of the limb, enabling high-acceleration movements without compromising kinematic complexity.

Notably, organisms utilize this seventh DoF—typically the knee joint between the femur and tibia—for critical posture regulation.
While six DoFs are kinematically sufficient to define a spatial pose, this redundant seventh DoF allows the organism to adjust limb posture without changing foot position.
The core significance of this mechanism lies in Force Transmission Optimization, enabling the organism to locate the optimal leverage point and maximize actuation efficiency across varying terrains \cite{haldane_robotic_2016}.

Currently, existing small-scale multi-legged robots generally exhibit limited mobility in complex environments, often attributed to insufficient drive power or inadequate DoFs.
\begin{enumerate}
    \item Integration of Servo Systems: Integrating compact servo actuators into miniaturized leg designs is challenging due to spatial constraints.
    \item Limitations of Serial Legs: Serial leg configurations increase the inertia of the end-effector, reducing dynamic performance and response speed.
    \item Complexity of High-DoF Designs: The design and integration of high-DoF leg mechanisms are inherently complex, making it difficult to ensure stability and reliability \cite{sreetharan_monolithic_2012}.
\end{enumerate}


To address these challenges, this paper presents a novel spatial four-degree-of-freedom (DoF) parallel leg mechanism, as shown in Fig.~\ref{fig:ant_real}.
The goal is to design a lightweight, miniaturized, high-DoF parallel mechanism and to evaluate how additional active DoFs enhance robot functionality.
The design aims to replicate both the kinematic agility and low-inertia characteristics of biological limbs through the following strategies:
\begin{enumerate}
    \item Structural Configuration and Integrated Layout: We adopt the biological kinematic topology of three DoFs at the hip and one DoF at the knee. To accommodate symmetrical actuator layout, all drive units—including the actuator for knee motion—are centrally integrated within the robot's thorax. This body-integrated design not only overcomes spatial constraints in servo integration but also enables omnidirectional attitude control of the two-link system while minimizing rotational inertia at the distal end.
    \item Dynamic Function and Significance: Unlike the conventional view that three DoFs are sufficient for centroidal control, the fourth DoF introduced here—which governs the inclination of the leg plane relative to the ground—goes beyond simple kinematic positioning. Instead, it is used to optimize force transmission in the leg linkage. By decoupling limb posture from contact force vectors, this redundant DoF significantly enhances locomotion stability and adaptability in unstructured environments.
\end{enumerate}

In summary, the main contributions of this work are as follows:
\begin{enumerate}
    \item Novel Parallel Mechanism Based on Merged Spherical Centers: We propose a spatial four-DoF parallel leg mechanism using a Spatial Four-Bar (SFB) linkage \cite{mcclintock_millidelta_2018}. By merging the rotational centers of the SFB joints, we simplify the complex spatial constraints into an equivalent parallel four-bar linkage. This design ensures structural symmetry and enables high-DoF spatial posture adjustment via merged output along the central axis.
    \item Body-Integrated Actuation and Low Inertia: With the proposed configuration, all actuators are fully integrated within the thorax and transmit motion via parallel linkages. This layout not only addresses spatial constraints in micro-robots but also significantly reduces the rotational inertia of the leg tip, enhancing dynamic performance. Moreover, the merged-center design yields a simplified kinematic model suitable for real-time control.
\end{enumerate}

The remainder of this paper is organized as follows: Section II presents the parallel design concept and circuit control design of the leg. Section III provides an analysis of the leg's kinematics and workspace. Section IV presents experimental results demonstrating the leg's motion capability and force output. Finally, the paper concludes with a summary of the proposed design and directions for future research.


%% file: 02_design.tex
\begin{figure*}
    \centering
    \includegraphics[width=1\linewidth]{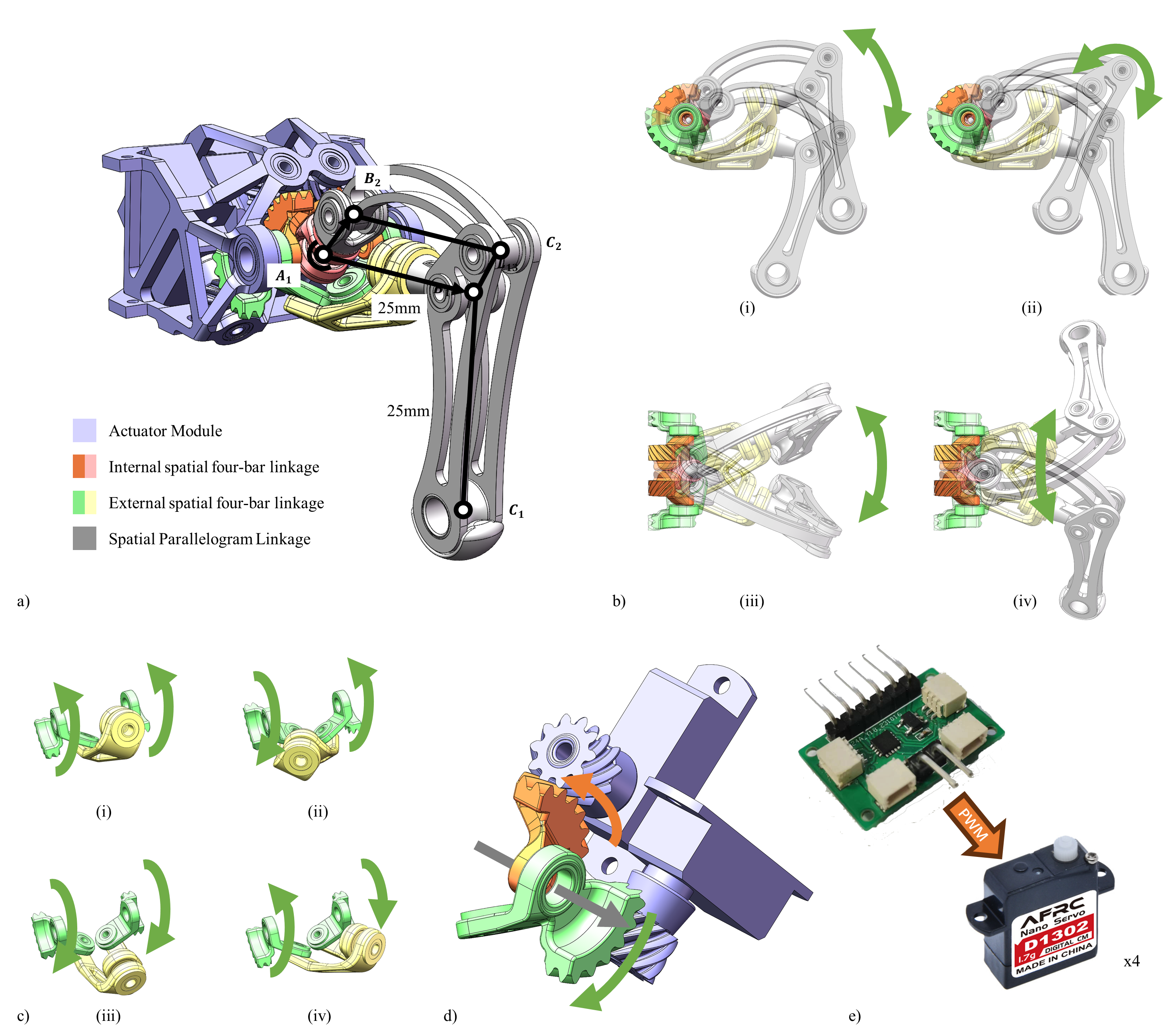}
    
    \caption{\textbf{Schematic illustration of the designed mechanism.} (a) overall view of the mechanism in space; (b) Kinematic schematic of the A$_1$B$_2$ linkage; (c) Kinematic schematic of the spherical linkage; (d) Drive transmission configuration of the mechanism; (e) The drive and control circuit.}
    \label{fig:leg_design}
\end{figure*}

\section{Design}

This work presents an innovative spatial four-bar linkage design for the leg-foot system of a micro robot.
Conventional designs typically use planar parallel four-bar linkages.
While these can achieve two-degree-of-freedom (2-DoF) motion and centralize the drive motors at the base to reduce moving inertia, their motion is confined to a single plane.
To enhance the locomotion agility of the leg-foot system, this work extends the planar four-bar linkage to a spatial configuration.

As shown in Fig.~\ref{fig:leg_design}, a single leg comprises a power module, two spherical five-bar linkages, and one spatial four-bar linkage.
By upgrading the planar four-bar joint to a spatial configuration, the two spherical five-bar linkages are responsible for integrating four motion inputs and transmitting the synthesized motion to the spatial four-bar linkage.
This structure allows the links to deviate from their original working plane, thereby endowing the four-bar leg with additional capabilities of internal rotation and abduction.

Fig.~\ref{fig:leg_design} illustrates the pose characteristics of the spatial parallel four-bar mechanism, which are governed by the spatial orientation of links $A_1B_1$ and $A_1B_2$. As shown in Fig.~\ref{fig:leg_design}(a), in-phase pitching of $A_1B_1$ and $A_1B_2$ generates hip pitching motion. Out-of-phase pitching, depicted in Fig.~\ref{fig:leg_design}(b), results in knee flexion/extension. In-phase abduction, as in Fig.~\ref{fig:leg_design}(c), produces hip abduction, while out-of-phase abduction (Fig.~\ref{fig:leg_design}(d)) induces hip internal rotation. Note that the system approaches a kinematic singularity when $A_1B_1$ and $A_1B_2$ become nearly collinear, leading to loss of actuation authority around the third rotational axis.

To achieve active two-degree-of-freedom (DoF) control of the link, the design synthesizes base motion into spherical motion using a spherical five-bar linkage.
As shown in Fig. 4, the topology of the spherical five-bar joint is configured to satisfy the kinematic requirements of the spatial four-bar linkage.
This mechanism defines the screw relationship between the input (driving) and output (driven) axes.
The two input axes are arranged coaxially, decoupling pitch motion—i.e., motion that induces only overall pitching of the linkage without exciting other DoFs.
All four input axes adopt this coaxial layout, which significantly simplifies the transmission system.
Pure pitch motion is produced when the dual input axes move in phase; a phase difference generates lateral deflection.
In addition to the two DoFs controlled by the dual input axes, the output axis of the spherical five-bar joint has an unconstrained rotational freedom along its own axis.
If the two spherical five-bar joints are not concentric, additional revolute joints would be needed to compensate for motion deviation, substantially increasing mechanism complexity.
Therefore, the design adopts a nested spherical mechanism: the inner spherical five-bar linkage uses a compact topology, allowing the outer layer to achieve a fully enveloping concentric configuration.

The motion output of this configuration is currently constrained by the servo motors' stroke limits, and thus does not fully realize its design potential.
The effective working length of the leg is set to 25 mm with an initial joint angle configuration of 45°.

The drive system consists of four servo motors arranged in an array within the drive module, connected to a parallel transmission architecture via helical gear sets.
The main body of each leg is fabricated using high-precision additive manufacturing.
Key load-bearing links are CNC-machined from carbon fiber reinforced polymer (CFRP) and bonded with epoxy structural adhesive.
The foot tip is equipped with a high-damping silicone friction pad to optimize dynamic contact behavior.
Each leg unit is driven by a dedicated control board featuring an STM32C011 microcontroller, which generates four-channel PWM signals to actuate the servo motors.

%% file: 03_analysis.tex
\section{Analysis}

\input{03_analysis_kinematics}

\input{03_analysis_SFB}

\input{03_analysis_workspace}

%% file: 03_analysis_kinematics.tex
\subsection{Spatial four-bar linkage}
\begin{figure}
    \centering
    \includegraphics[width=1\linewidth]{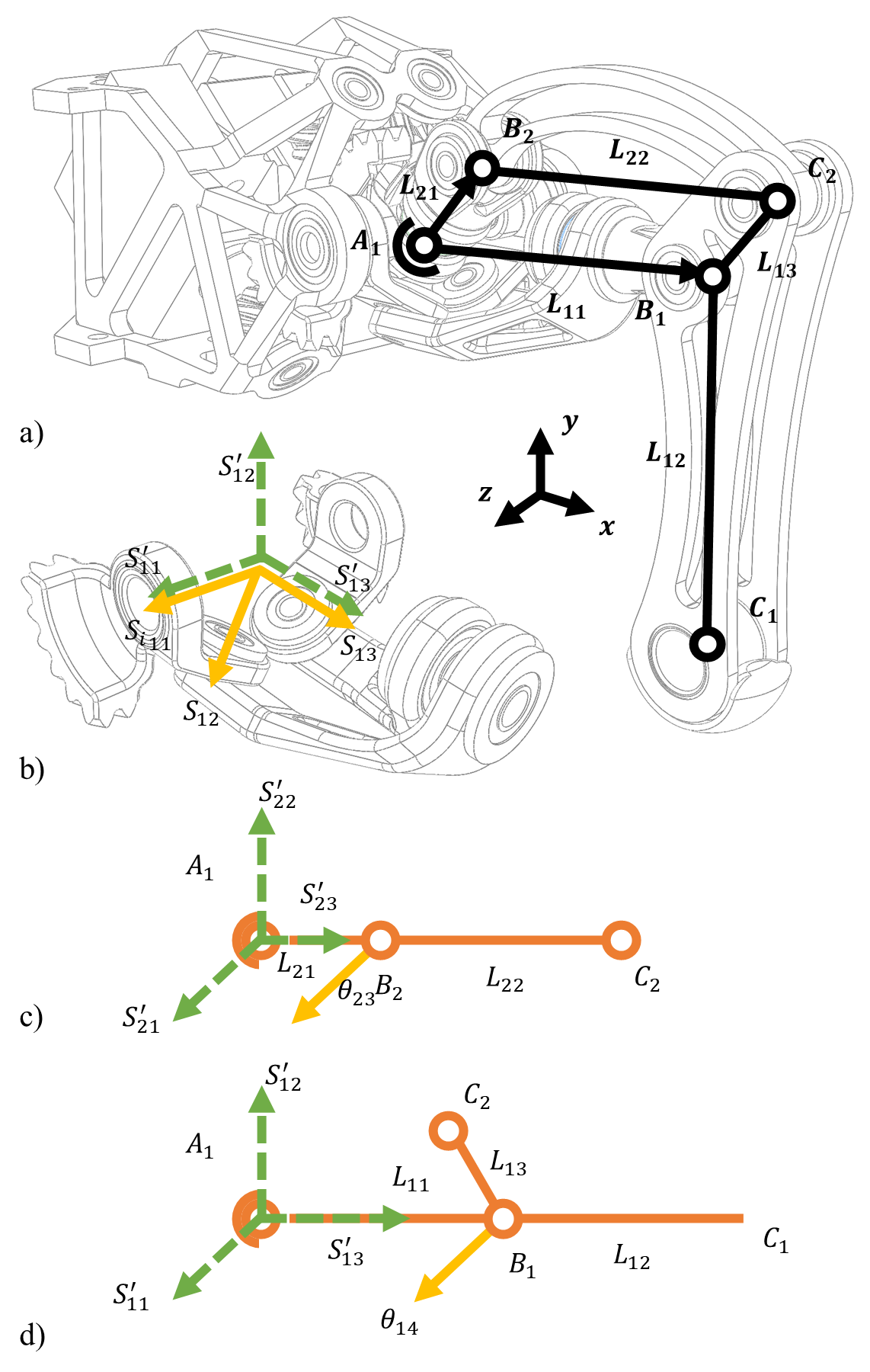}
    \caption{Kinematic Modeling of Spatial Four-bar Linkage and Spherical Five-bar Linkage}
    \label{fig:kinematics}
\end{figure}

As shown in Fig. \ref{fig:kinematics}, the leg kinematic chain of the hexapod robot essentially forms a spatial four-bar mechanism. Due to the non-orthogonal joint configuration of the previously designed SFB joint, the complexity of kinematic modeling is significantly increased. To simplify the kinematic analysis, a virtual orthogonal joint coordinate system is established at the center of the SFB joint, with its spatial configuration illustrated in Fig. \ref{fig:kinematics}.

From the kinematic structure shown in Fig. \ref{fig:kinematics}, the virtual leg kinematic chain can be derived.
\begin{equation}
T_{C_1}(\theta) = e^{[S_{11}'] \theta_{11}'} e^{[S_{12}'] \theta_{12}'} e^{[S_{13}'] \theta_{13}'} e^{[S_{14}] \theta_{14}} M_{C_1}
\label{eq:kine_01}
\end{equation}

The virtual joint angles can be obtained by solving the forward kinematic equations.
\begin{equation}
\begin{aligned}
\begin{cases}
X_{C_1}^2 + Y_{C_1}^2 + Z_{C_1}^2 = L_{11}'^2 + L_{12}'^2 - 2 L_{11}' L_{12}' \cos(\theta_{14}), \\
- s_{12}' (L_{11}' + L_{12}' c_{14}) + c_{12}' (L_2 s_{13}' s_{14}) = Z_{C_1}, \\
- s_{11}' (L_2 c_{13}' s_{14}) + c_{11}' \big( c_{12}' (L_{11}' + L_{12}' c_{14}) + s_{12}' L_2 s_{13}' s_{14} \big) = X_{C_1}.
\end{cases}
\end{aligned}
\label{eq:kine_02}
\end{equation}

Considering that the leg of the hexapod robot forms a spatial parallelogram mechanism, the azimuth angle of link $B_1C_2$ remains synchronized with that of link $A_1B_2$, resulting in the following constraint relationship:
\begin{equation}
e^{[S_{21}'] \theta_{21}'} e^{[S_{22}'] \theta_{22}'} e^{[S_{23}'] \theta_{23}'} = e^{[S_{11}'] \theta_{11}'} e^{[S_{12}'] \theta_{12}'} e^{[S_{13}'] \theta_{13}'} e^{[S_{13}'] (\theta_{14} + \frac{3\pi}{4})}
\label{eq:kine_03}
\end{equation}

By solving the Eq. \eqref{eq:kine_03}, the virtual joint coordinates of the second SFB joint can be derived:
\begin{equation}
\begin{aligned}
\begin{cases}
\theta_{22}' = \arcsin\left( s_{12}' c_{14 + \frac{3\pi}{4}} - c_{12}' s_{13}' s_{14 + \frac{3\pi}{4}}\right) \\
\theta_{21}' = \theta_{11}'-\arctan\left(\frac{c_{13}' s_{14 + \frac{3\pi}{4}}}{c_{12}' c_{14 + \frac{3\pi}{4}} + s_{12}' s_{13}' s_{14 + \frac{3\pi}{4}}}\right)
\end{cases}
\end{aligned}
\label{eq:kine_04}
\end{equation}

%% file: 03_analysis_SFB.tex
\subsection{Five-bar spherical linkage}

After obtaining the virtual joint angles, they are used to calculate the input angles of the SFB joint. The output azimuth angle of the SFB joint is the same as the azimuth angles of the first three virtual joints, thus we can obtain: 
\begin{equation}
e^{[S_{i_{11}}] \theta_{i_{11}}} e^{[S_{12}] \theta_{12}} e^{[S_{13}] \theta_{13}} = e^{[S_{11}'] \theta_{11}'} e^{[S_{12}'] \theta_{12}'} e^{[S_{13}'] \theta_{13}'}
\label{eq:SFB_01}
\end{equation}

Expanding this, we get:
\begin{equation}
\begin{cases}
c_{11}' c_{12}' = c_{i_{11}} c_{12} - \frac{1}{2} s_{i_{11}} s_{12} \\
s_{11}' c_{12}' = s_{i_{11}} c_{12} + \frac{1}{2} c_{i_{11}} s_{12} \\
\frac{\sqrt{3}}{2} s_{12} = - s_{11}'
\end{cases}
\label{eq:SFB_02}
\end{equation}

Simplifying, we obtain:
\begin{equation}
\begin{cases}
\theta_{i_{11}} = \theta_{11}' - \arctan\left( \frac{1}{2} t_{12} \right) \\
\theta_{i_{12}} = \theta_{11}' + \arctan\left( \frac{1}{2} t_{12} \right)
\end{cases}
\label{eq:SFB_03}
\end{equation}

Where:
\begin{equation}
\theta_{12} = \arcsin\left( \frac{2\sqrt{3}}{3} s_{12}' \right) \\
\label{eq:SFB_04}
\end{equation}

%% file: 03_analysis_workspace.tex
\subsection{Workspace}

\begin{figure}
    \centering
    \includegraphics[width=1\linewidth]{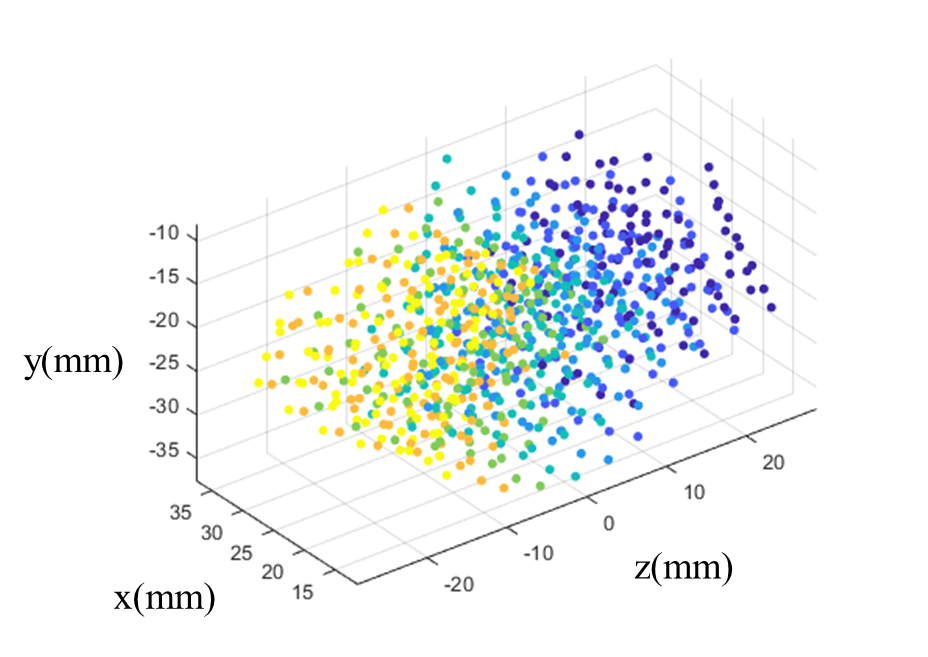}
    \caption{Workspace of the leg}
    \label{fig:workspace}
\end{figure}

Fig.~\ref{fig:workspace} illustrates the workspace of the proposed leg, with color mapping indicating different values of \(\theta_3\) to highlight its effect—a key distinction from other bioinspired leg designs. When \(\theta_3\) is fixed, the workspace covers approximately 7681\(\,\)mm\(^3\). In contrast, enabling \(\theta_3\) expands the workspace to about 22255\(\,\)mm\(^3\).


%% file: 04_experiment.tex
\section{Experiment}



\begin{figure}
    \centering
    \includegraphics[width=1\linewidth]{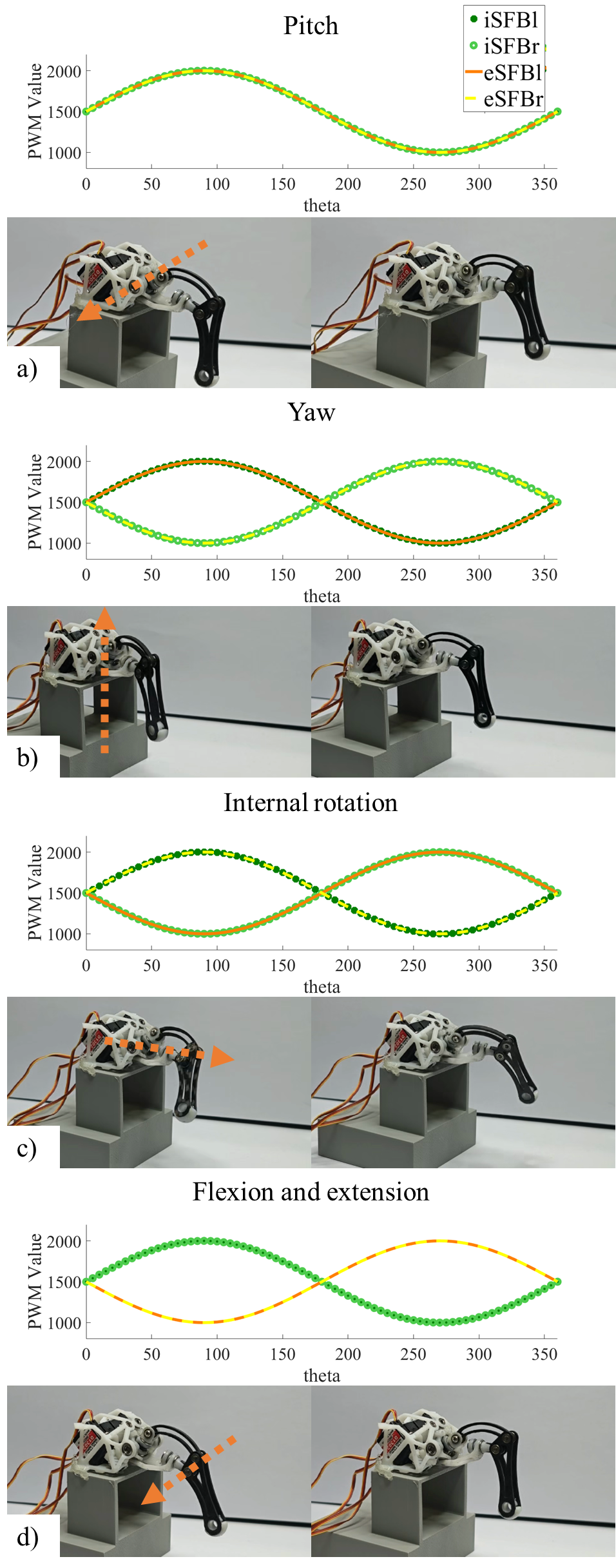}
    \caption{Motion demonstration of four input channels with out-of-phase motion inputs}
    \label{fig:experiment01}
\end{figure}

\subsection{Workspace Demonstration}


To illustrate the workspace of the parallel leg, four phase combinations of sinusoidal input signals are applied, as shown in Fig.~\ref{fig:experiment01}:
\begin{itemize}
\item \textbf{Phase 1:} Channels 1 and 3 lead Channels 2 and 4 by $\pi$. No relative motion occurs inside the five-bar linkage; the end-foot moves in the YZ plane.
\item \textbf{Phase 2:} Channels 1 and 2 lead Channels 3 and 4 by $\pi$. The internal motions of the five-bar linkage are opposite; the end-foot moves in the XY plane.
\item \textbf{Phase 3:} Channels 1 and 3 are in phase with Channels 2 and 4. The internal motions are opposite; the end-foot moves in the XZ plane.
\item \textbf{Phase 4:} Channels 1 and 3 lag Channels 2 and 4 by $\pi$. The internal motions are in the same direction; the end-foot moves in the XZ plane.
\end{itemize}
Thus, the workspace of the parallel leg approximates a spherical surface.

\subsection{Inverse Kinematics Trajectory Demonstration}

\begin{figure}
    \centering
    \includegraphics[width=1\linewidth]{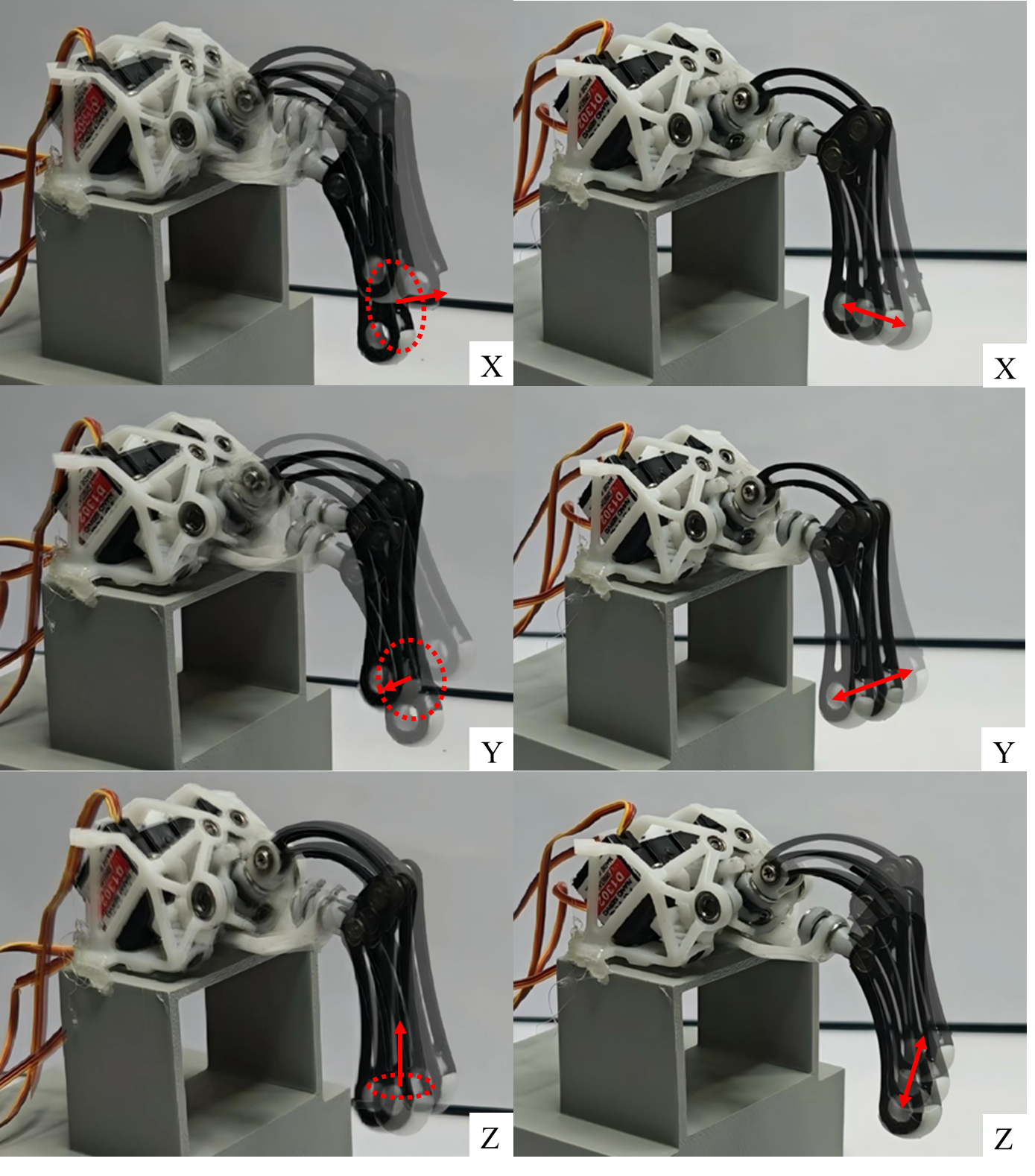}
    \caption{Physical demonstration of the workspace of the leg}
    \label{fig:experiment02}
\end{figure}


The end-foot was commanded to follow the green trajectory shown in Fig.~\ref{fig:experiment02} in the XY, XZ, and YZ planes. Additionally, linear motions along the X, Y, and Z axes were performed separately.

\subsection{Force Output Demonstration}

\begin{figure}
    \centering
    \includegraphics[width=1\linewidth]{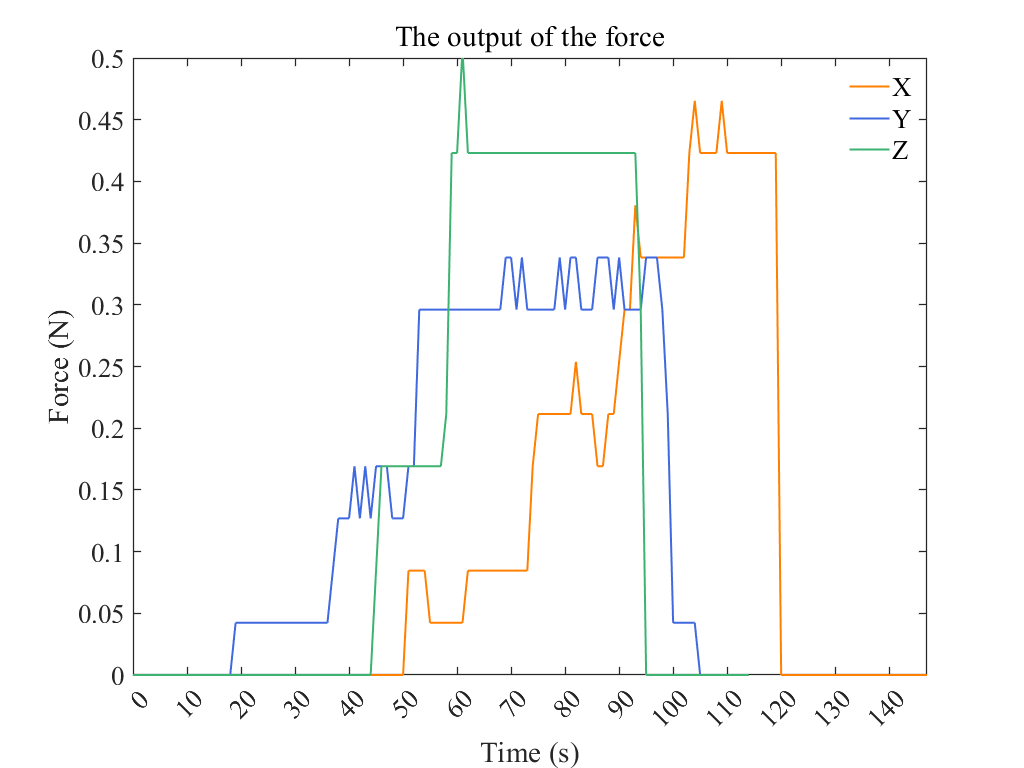}
    \caption{Force output in three directions}
    \label{fig:force_plot}
\end{figure}

Following the linear trajectory described in Part B (using the Z-direction experiment as an example), the end-foot was actuated to apply force to the strain gauge sensor. The resulting output waveform is shown in Fig.~\ref{fig:force_plot}, with a peak output force of approximately 0.5 N.

%% file: 05_conclusion.tex
\section{Conclusion}
In this work, we developed a micro-scale parallel biomimetic leg with four degrees of freedom, addressing the inherent limitations in motion often encountered in traditional micro-robots. The design integrates all four actuators within the robot's body, synthesizing hip motion with three degrees of freedom and knee motion with one degree of freedom through parallel kinematics. The total mass of the leg system is only 18.9 g, with an end-effector output force of approximately 0.5 N and a workspace exceeding 22255 mm³. This design enables micro-scale robots to achieve more agile and versatile locomotion capabilities.

Our future work will focus on further miniaturizing the parallel leg mechanism, including the incorporation of origami-inspired structures and micro-actuators. Additionally, we aim to integrate multiple copies of the proposed leg module onto a single robotic platform to achieve enhanced actuation and mobility for micro-robots.

%% file: 06_reference.tex
\bibliographystyle{IEEEtran}
\bibliography{references}